\documentclass[letterpaper]{article} 
\usepackage[]{aaai23}  
\usepackage{times}  
\usepackage{helvet}  
\usepackage{courier}  
\usepackage[hyphens]{url}  
\usepackage{graphicx} 
\usepackage{natbib}  
\usepackage{caption} 
\usepackage{algorithm}
\usepackage{algorithmic}

\usepackage{comment}
\usepackage{dirtytalk}

\usepackage[textwidth=15mm,disable]{todonotes}

\usepackage{siunitx}
\usepackage{subfig}
\usepackage{tikz}
\usepackage{textcomp}
\usepackage{tikz}
\usepackage{booktabs}

\usetikzlibrary{calc,patterns}
\usetikzlibrary{positioning}
\usetikzlibrary{matrix}
\usepackage{newfloat}
\usepackage{listings}
\DeclareCaptionStyle{ruled}{labelfont=normalfont,labelsep=colon,strut=off} 
\floatstyle{ruled}
\newfloat{listing}{tb}{lst}{}
\floatname{listing}{Listing}
\title{Machine Learning Meets The Herbrand Universe}
\author{
    Jelle Piepenbrock, \textsuperscript{\rm 1, 2} Josef Urban, \textsuperscript{\rm 2}  Konstantin Korovin, \textsuperscript{\rm 3} Miroslav Olšák, \textsuperscript{\rm 4} Tom Heskes \textsuperscript{\rm 1} and Mikolaš Janota \textsuperscript{\rm 2}
}
\affiliations{
    \textsuperscript{\rm 1}Radboud University Nijmegen, the Netherlands\\
    \textsuperscript{\rm 2}Czech Technical University in Prague, Czech Republic\\
    \textsuperscript{\rm 3}University of Manchester, United Kingdom\\
    \textsuperscript{\rm 4}Institut des Hautes Études Scientifiques Paris, France\\
    
}

\usepackage{bibentry}

\begin{document}

\maketitle

\begin{abstract}
  The appearance of strong CDCL-based propositional (SAT) solvers has
  greatly advanced several areas of automated reasoning (AR). One of
  the directions in AR is thus to apply SAT solvers to expressive
  formalisms such as first-order logic, for which large corpora of
  general mathematical problems exist today. This is possible due to
  Herbrand's theorem, which allows reduction of first-order problems
  to propositional problems by instantiation. 
  The core
  challenge is choosing the right instances from the typically
  infinite Herbrand universe.

  In this work, we develop the first
  machine learning system targeting this task, addressing its
  combinatorial and invariance properties. In particular, we develop a
  GNN2RNN architecture based on an invariant graph neural network
  (GNN) that learns from problems and their solutions independently of
  symbol names (addressing the abundance of skolems), combined with a
  recurrent neural network (RNN) that proposes for each clause its
  instantiations.  The architecture is then trained on a corpus of
  mathematical problems and 
  their instantiation-based
  proofs, and its performance is evaluated in several ways. We show
  that the trained system achieves high accuracy in predicting the
  right instances, and that it is capable of solving many 
  problems by educated guessing when combined with a ground solver. To
  our knowledge, this is the first convincing use of machine learning
  in synthesizing relevant elements from arbitrary Herbrand universes.
\end{abstract}

\section{Introduction}
\label{sec:intro}


Quantifiers lie at the heart of mathematical logic, modern mathematics and reasoning. They enable
expressing statements 
about infinite domains. Practically all today's systems used for
formalization of mathematics and software verification are based on
expressive foundations such as first-order and higher-order logic, set
theory and type theory, that make essential use of quantification.

Instantiation is a powerful tool 
for formal reasoning with quantifiers. 
The power of instantiation is formalized by \emph{Herbrand's theorem}~\cite{herbrand},
which states, roughly speaking, that  within  first-order logic (FOL),
quantifiers can always be eliminated by the right instantiations.
Herbrand's theorem further states that  it is sufficient to consider instantiations from the \emph{Herbrand universe},
which 
consists of
terms with no variables (\emph{ground terms}) constructed from the symbols appearing in the
problem. 
This 
fundamental
result has been explored in automated reasoning (AR) systems since the 1950s~\cite{Davis01}.
In particular, once the right instantiations are discovered,
the problem
typically becomes easy to decide 
by methods based on state-of-the-art SAT solvers~\cite{handbook-sat}.

Coming up with the right instantiations is however 
nontrivial.
The space of possible instantiations (the Herbrand universe) is typically infinite  and complex.
The general undecidability of 
theorem proving is obviously
connected to the hardness of finding 
the right instantiations, which includes finding arbitrarily complex mathematical objects.

\textbf{Contributions:} In this work we 
develop the first machine
learning (ML) methods that automatically propose suitable
instantiations. This is motivated both by the growing ability of ML
methods to prune the search space of
automated theorem provers (ATPs)~\cite{KaliszykUMO18}, 
and also by their growing
ability to synthesize various logical data 
\cite{Gauthier20,UrbanJ20}. 
In particular:
\begin{enumerate}
\item We construct an initial 
  corpus of instantiations by 
repeatedly running a randomized grounding procedure followed by a ground solver (Section~\ref{sec:solver}) on
113332 clausal ATP problems extracted from the Mizar Mathematical
Library. 
We analyze the solutions, showing that almost two thirds of
the instances contain newly introduced skolem symbols created by the
clausification (Section~\ref{sec:data}). 
\item  We develop a targeted GNN2RNN neural architecture based on
a graph neural network (GNN) that learns to characterize the problems
and their clauses independently of the symbol names (addressing the
abundance of skolems), combined with a recurrent neural network (RNN) that
proposes for each clause its instantiations based on the GNN
characterization (Section~\ref{sec:gnn}). 
\item  The GNN2RNN is 
  trained and used 
  to propose instances for the problems. Its training starts with the randomized solutions and continues 
  by learning from its own successful predictions.
We show that the system achieves high accuracy in predicting the instances (Section~\ref{sec:exp}).
\item Finally, the trained neural network when combined with the
  ground solver 
  is shown to be able to solve a large number of
  testing problems by educated guessing (Section~\ref{sec:exp}). To our knowledge, this is the first convincing use of machine learning
  in general synthesis of relevant elements from arbitrary Herbrand universes.
\end{enumerate}

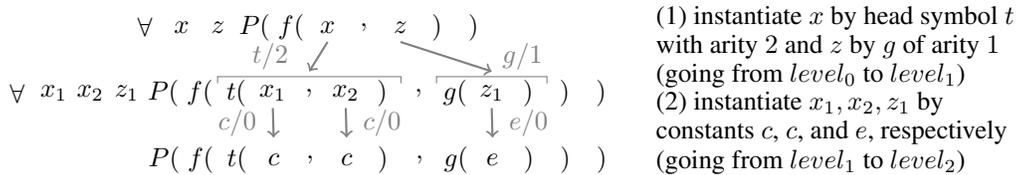
\begin{figure*}[t]
  \centering
\begin{tikzpicture}[yscale=0.88,inner sep=1pt]
  \node[align=left,right] at (13em, 1.7) {(1) instantiate $x$ by head symbol $t$\\with arity $2$ and $z$ by $g$ of arity $1$\\ (going from $level_0$ to $level_1$)};
  \node[align=left,right] at (13em, 0.4) {(2) instantiate $x_1,x_2,z_1$ by\\constants $c$, $c$, and $e$, respectively\\ (going from $level_1$ to $level_2$)};
	\matrix [matrix of math nodes,nodes={text width=1.18em,align=center,minimum height=1.1em,anchor=center}] (A) at (0, 2)
	{\forall& x & z  & P(  & f( & x &, & z & )  & ) \\ };
	\matrix [matrix of math nodes,nodes={text width=1.18em,align=center,minimum height=1.1em,anchor=center}] (B) at (0, 1)
  {\forall  & x_1 & x_2 & z_1  & P(  & f( & t( &x_1 &, & x_2&) &, &g( & z_1 &) & ) & )  \\ };
	\matrix [matrix of math nodes,nodes={text width=1.18em,align=center,minimum height=1.1em,anchor=center},right] (C) at (-6.3em, 0)
  {P(  &f( & t( &c &, & c&) &, &g( & e &) & ) & )  \\ };

  \draw[thick,gray,->] (A-1-6.south) to node[left=1em]{$t/2$} ( $(B-1-9.north) + (0,0.1)$ );
  \draw[thin,gray] ( $(B-1-7.north west) - (0,0.1)$ ) -- (B-1-7.north west) -- (B-1-11.north east) -- ( $(B-1-11.north east) - (0,0.1)$ );

  \draw[thick,gray,->] (A-1-8.south) to node[right=20pt]{$g/1$} ( $(B-1-14.north) + (0,0.1)$ );
  \draw[thin,gray] ( $(B-1-13.north west) - (0,0.1)$ ) -- (B-1-13.north west) -- (B-1-15.north east) -- ( $(B-1-15.north east) - (0,0.1)$ );

  \draw[thick,gray,->] (B-1-8.south) to node[left=5pt]{$c/0$} ( $(C-1-4.north) + (0,0.1)$ );
  \draw[thick,gray,->] (B-1-10.south) to node[right=5pt]{$c/0$} ( $(C-1-6.north) + (0,0.1)$ );
  \draw[thick,gray,->] (B-1-14.south) to node[right=5pt]{$e/0$} ( $(C-1-10.north) + (0,0.1)$ );
\end{tikzpicture}
  \caption{Term instantiation through incremental deepening. In the
    figure, there are two  instantiation steps, one after
    the other.}\label{fig:deepening}
\end{figure*}

\section{Ground Solver}
\label{sec:solver}


There are several ways how to combine instantiation of clausal first-order 
problems with decidable and efficient (un)satisfiability checking of
their proposed ground instances. The most direct approach (used in
instantiation-based ATPs such as iProver~\cite{iprover}) is to
explicitly add axioms for equality, allowing their instantiation as for any other axioms, and
directly use SAT solvers for the ground checking. An alternative
approach is to avoid explicit addition of the equality axioms, and
instead use combinations of SAT solvers with ground congruence
closure~\cite{simplify,NieuwenhuisOliveras07}.

We have explored both approaches and ultimately decided to use the
latter in this work. The main reason is that the combinations of SAT
solvers with ground congruence closure are today very efficiently
implemented~\cite{handbook:smt}, posing practically no issues even with thousands of
instances. Using the most direct approach would, on the other hand,
require a large number of additional instances of the equality axioms
to successfully solve the ground problems. In our preliminary
measurements, the average ratio of such necessary additional instances
was over 40\%, which would exponentially decrease the chance of randomly predicting the right set of instances.

%

In more detail, we use as our ground solver an efficient combination
of a SAT solver with ground congruence closure provided in the Vampire
automated theorem prover~\cite{vampire}.  Here, the SAT solver
abstracts first-order logic atoms as propositional variables and
starts producing satisfying assignments (models) of this abstraction,
which are then checked against the properties of equality
(reflexivity, transitivity, congruence, symmetry). This process
terminates when a model is found that satisfies the equality
properties, or when the SAT solver runs out of models to try. In that
case, the original problem is unsatisfiable, which means that the
ground instances were proposed correctly by the ML system.

\section{Dataset of Mathematical Problems}
\label{sec:data}

We construct a corpus of instantiations
by repeatedly running a randomized grounding procedure (Section~\ref{sec:random})
on  $\num{113332}$ first-order ATP problems made available to us by the
AI4REASON project.\footnote{\url{https://github.com/ai4reason/ATP_Proofs}}
They originate from the Mizar Mathematical Library
(MML)~\cite{KaliszykU13b} and are exported to first-order logic by the
MPTP system~\cite{Urban06}.  All these problems have an ATP proof (in
general in a high time limit) found by either the
E/ENIGMA~\cite{Schulz13,anonenigma} or
Vampire/Deepire~\cite{vampire,deepire} systems. Additionally, the
problems' premises have been
\textit{pseudo-minimized}~\cite{holyhammer} by iterated Vampire
runs. We use the pseudo-minimized versions because our focus here is
on guiding instantiation rather than premise selection. The problems come from $\num{38108}$ \textit{problem
  families}, where each problem family corresponds to one original
Mizar theorem. Each of these 
theorems can have multiple minimized ATP proofs using different sets
of premises.
The problems range
from easier to challenging ones, across various mathematical fields
such as topology, set theory, logic, algebra and linear algebra, real,
complex and multivariate analysis, trigonometry, number and graph
theory, etc. 


We first clausify the problems by E, and then repeatedly run the randomized grounding procedure on all of
them, followed by the ground solver using a 30s time limit. The first run solves
3897 of the problems, growing to 7790 for the union of the first 9 runs, and to 11675 for the union of the first 100 runs.   The 113332 problems
have on average 35.6 input clauses and the 3897 problems solved in the first run have on average 12.8 input clauses.
%
%
6.0 instances are needed on average to solve a
problem.\footnote{Over 40 instances (max. 63) are used in
  some problems}

%
Note that each clause can in general be instantiated
more than once. 
Also, 3.9 (almost two thirds) of the instances contain
on average at least one skolem symbol.

This means that we strongly
need a learning architecture invariant under symbol renamings, rather
than off-the-shelf architectures (e.g., transformers) that depend on
fixed consistent naming. This motivates our use of a
property-invariant graph neural architecture (Section~\ref{sec:gnn}).

\subsection{Randomized Grounding Procedure}
\label{sec:random}
Randomized grounding can be parameterized in various ways. To develop
the initial dataset here we use a simple multi-pass randomized
grounding with settings that roughly correspond to our ML-guided
instantiation architecture (Section~\ref{sec:gnn}).  These settings
are as follows. We use at most two passes (levels) of instantiation for every
input clause.  In the first pass,  for each variable we randomly
select an arbitrary function symbol from the problem's signature, and
provide it (if non-constant) with fresh variables as arguments. In the
second pass, we ground all variables with randomly selected
constants from the problem's signature. The first pass is repeated 25 times for each clause in its
input, and the second pass 5 times. The input to the first pass is the
original clausal problem.  The input to the second pass is the
(deduplicated) union of the clauses produced in the first pass and of
the original clauses.

This means that each input clause can produce $(25+1)*5=130$ ground
instances, potentially resulting in ground problems with thousands of
ground clauses.  As noted in Section~\ref{sec:solver} such input sizes
typically pose no problems to the ground solver. The average ground
problem sizes are however typically below 1000, because of the
overlaps and limited number of options during the random grounding.

\section{The GNN2RNN Instantiation Architecture}
\label{sec:gnn}


Mathematical problems have often many symmetries, making them
challenging for naive use of off-the-shelf sequence-based learning
methods. Our clausal problems are invariant under reordering of
clauses and literals, renaming of variables in each clause, and also
under consistent renaming of symbols in a problem.  To address that,
we base our architecture on a graph neural network (GNN) with such properties proposed
by~\cite{piegnn} and used so far in several ATP
tasks~\cite{anonenigma,tpcomponents} to \emph{classify existing
objects}.  In this work, we reimplement the GNN in
PyTorch~\cite{NEURIPS2019_9015}, and add to it
a novel anonymous, signature-bound recurrent neural network (RNN)
that allows us to also \emph{generate new objects} (clause
instantiations). 
To our knowledge, this is the first ML architecture combining such
strong invariant and nameless problem encoding with the need for
non-anonymous symbolic decoding (synthesis).

\begin{figure}[]
  \centering
\begin{tikzpicture}[yscale=0.88,inner sep=1pt,every node/.style={align=left,right}]
  \matrix [matrix of nodes,row sep={2em,between origins},column sep=1em] (A)
  {
    $\forall xz.\,P(  f( x,  z )  )$ & $x:t$ & $z:g$ \\ 
    $\forall x_1x_2z_1.\, P(f(t(x_1,x_2),  g(z_1)))$ & $x_1:c$ & $x_2:c$ & $z:e$ \\ 
    $P(f(t(c,c), g(e)))$ &&&\\
  };
  \draw[thick,gray] (5,1.01) -- (5,-1.01);
  \draw[thick,gray,bend right,->] (A-2-1) edge node[below] {\footnotesize GNN} (A-2-2);
  \draw[thick,gray,bend right,->] (A-2-2) edge node[below] {\footnotesize RNN} (A-2-3);
  \draw[thick,gray,bend right,->] (A-2-3) edge node[below] {\footnotesize RNN} (A-2-4);
\end{tikzpicture}
  \caption{Predictions  corresponding to Figure~\ref{fig:deepening}. A GNN
  communicates the formula to the predictor on each line and previous
predictions are communicated by a hidden state.}\label{fig:typewriter}
\end{figure}
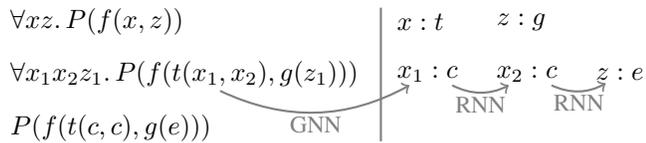

We first give a high-level view of how the network predicts the instantiations.
The
prediction is decomposed into levels,  where each level  deepens the predicted
terms.  The idea is depicted in Figure~\ref{fig:deepening}.  In each level, the
network  predicts a single function symbol  for each  variable of a given clause $C$.  Then,  a
new instance $C_1$ of $C$ is created by replacing the variables with the proposed function symbols and fresh variables as their arguments.
This whole process is iterated.  This means that the
network  never explicitly  sees that the terms are being deepened because it
predicts one deepening  step  at a  time.

Besides the increasing depth, the network also needs to be able to deal with
an arbitrary number of variables in each clause. This is handled in
an RNN fashion---variables are being predicted in a fixed order and the
information about the previous predictions is stored in a hidden state.
Overall, we can imagine the network prediction process as a typewriter, where
each line corresponds to deepening a level and columns correspond to
predicted symbols. This is illustrated by Figure~\ref{fig:typewriter}.
Note that each line may have different length---the length of the line  increases or 
decreases depending of the arity of the predicted symbols.  In particular, if all
predicted symbols are constants (symbols of arity 0),  the line is empty and the process stops. While the example discussed in Figure~\ref{fig:typewriter} is a valid example, in practice there is more complexity that needs to be dealt with. We are most often required to predict instances for multiple clauses at the same time, as well as in some cases multiple instances for the same clause.  However, we do limit the complexity by only running the procedure for 2 iterations, meaning that variables can only get instantiated with terms of depth 2 or less.

The following text details the neural network architecture.


\paragraph{\textbf{Graph Neural Network:}}
The specific GNN architecture we use is specifically constructed to be invariant to
symbol names, as it treats the input clauses in a fully anonymous
manner~\cite{piegnn}. In addition, the network has a notion of negation. The
specific structure of clausified first-order logic problems is taken
into account, with clause nodes being able to communicate with
literals, terms being able to communicate with their subterms, and all
symbols being able to directly communicate with all terms they are
in. The graph
neural network is invariant to clause order permutation and literal
permutation.
For our purposes here, it is enough that after several message passing rounds, the GNN outputs
vector representations for the three types of nodes in the graph: the
\textit{term nodes} \textbf{T}, the \textit{symbol nodes} \textbf{S}
and the \textit{clause nodes} \textbf{C}.
\begin{figure*}[t]
    \centering
    \includegraphics[width=0.7\textwidth]{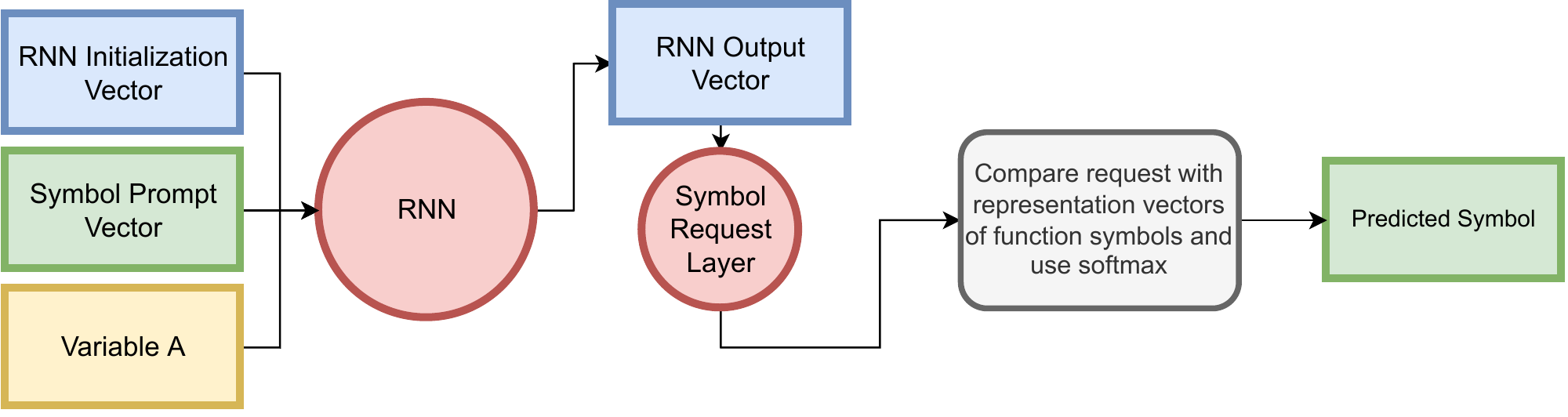}
    \caption{RNN architecture, shown predicting a symbol for Variable A, the first variable in a clause. A special, trainable \textit{symbol prompt vector} is used to mark a step where the RNN must predict a symbol for the queried variable.}\label{fig:rnnfig}
\end{figure*}

\paragraph{\textbf{RNN Function Symbol Prediction:}}
We modify the original GNN architecture to allow the network to
produce instantiations for each clause by using a recurrent
neural network (RNN) after running the GNN.
The setup is as follows. We predict an \textit{output vector} using
the RNN by taking for every clause the representation of the first
variable that occurs (in de Bruijn order) from the set \textbf{T}, a
special \textit{symbol prompt vector} and an initialization
vector.

This RNN output vector is processed by another layer to give
\textit{symbol request} vector and we compute the dot product between
this request vector and the representations of each function symbol in
the signature. We then apply the softmax function to get a probability
distribution over the function symbols for that variable (Figure
\ref{fig:rnnfig}). During training, we maximize the probability of
choosing the symbol used in the known proof.  During evaluation
(Section~\ref{sec:exp}) we can either (i) \emph{decode greedily},
choosing the maximum probability symbol, (ii) \emph{sample symbols}
according to the distribution defined by the model, or (iii) use a \textit{beam search} procedure to find the most likely sequences. In this work, we always use the sampling method (ii). We then continue
with the next step of the procedure (see Appendix, Figure
\ref{fig:rnnfig_cond}), where the RNN gets its own output from the
previous step, the chosen symbols, as well as the representation of
the second variable, etc.

This procedure preserves the anonymity of the entire
setup: the system is invariant to naming. In the end, we obtain a
mapping of variables to symbols for each clause. In addition, the
network can choose a special symbol when shown the first variable,
which indicates that the clause should not be instantiated. This means
the RNN predictor is simultaneously performing a
premise-selection-like task, while continuing to an instantiation task
when the clause is selected.


\textbf{Conditional Prediction:} To create symbol predictions that are conditioned on the symbols that were already chosen for other variables in the current clause, we created the setup as shown in Figure~\ref{fig:rnnfig_cond}. There, we schematically show the network in the process of predicting a symbol for the second variable (B) in a clause. First, the network get as its input an initialization vector, a variable representation vector and the representation vector of the symbol that was already chosen ($esk1\_0$) for that variable (A). The RNN then produces an output vector that should encode all the information about these prior decisions that is necessary to predict the next symbol. In the second step, the RNN processes this output vector, a vector representation of the second variable and a special \textit{symbol prompt vector} that indicates that the loading of previous decisions has ended and that we are currently expecting a new symbol prediction. The RNN then produces a new output vector, which we process as a symbol request using the \textit{symbol request layer}. The resulting vector is compared via a dot product with the vector representation of all the function symbols in the signature, giving a single float number for each comparison. This list of floats is converted into a probability distribution using the softmax function.

\begin{figure*}
    \centering
    \includegraphics[width=0.7\textwidth]{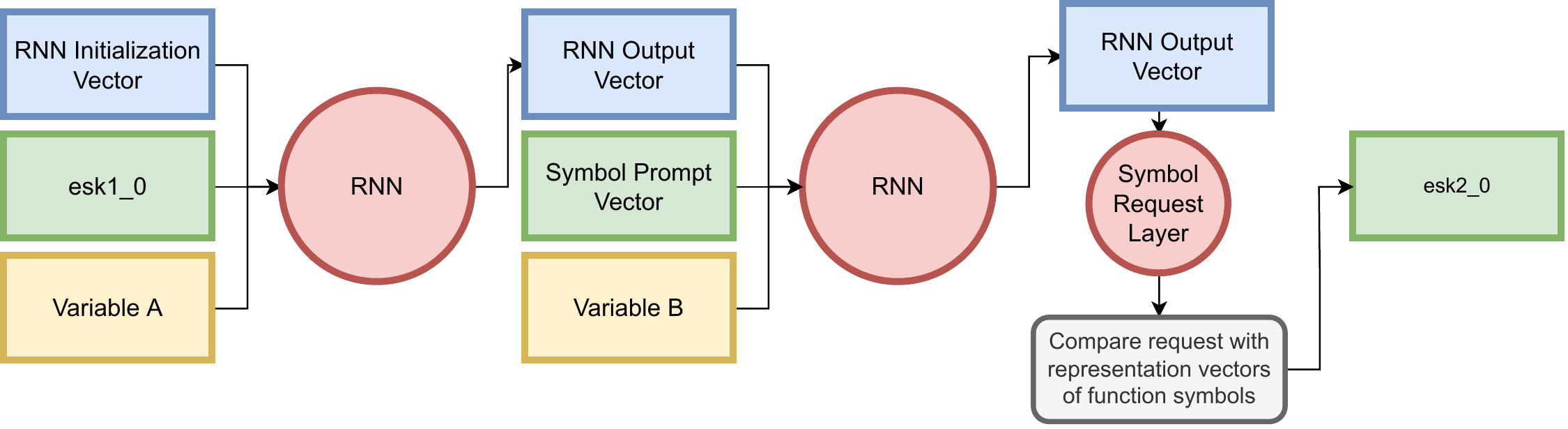}
    \caption{RNN architecture, shown in the process of predicting a symbol for the second variable B in a clause. $esk1\_0$ and $esk2\_0$ are two symbol node representations (for skolem constants). Here $esk1\_0$ was chosen for Variable A. Before using the \textit{symbol prompt vector}, the prefix of currently assigned symbols is shown to the RNN and the RNN can predict a symbol for variable B conditioned on the choice made for A.}
    \label{fig:rnnfig_cond}
\end{figure*}

\section{Experiments}
\label{sec:exp}
\todo{explain why and how only 2 levels are used}
\subsection{Datasets: Full and M2K Subset}
While we have 113332 problems in the full dataset, a smaller subset was selected to allow quicker iteration of experiments. The problems selected correspond to 2003 Mizar theorems known as the M2K subset, which is a subset of related Mizar articles~\cite{KaliszykUMO18}. Since we typically have multiple premise selections proving the same Mizar theorem, 4817 problems constitute the dataset which we will refer to as our \textit{M2K Dataset}. For the machine learning experiments, the Mizar theorems were split into 90\% training and development data and 10\% testing data. 5\% of the training and development data was used as the development set and 95\% as training data. Note that our split keeps problems which are versions of the same theorem in the same section of the split, so that data leakage between minor variations of the same proof idea is prevented. Also, the fact that a problem is assigned to the training set does not imply we already have a proof: the split is done independently of the availability of a solution. This means that each of these sets is a mixture of problems that already have the proof and problems that do not have one.

\subsection{Hardware}
All experiments were run either on a DGX machine with 8 NVIDIA Tesla V100 GPUs with 32GB memory, 512GB RAM and 80 cores of Intel(R) Xeon(R) CPU E5-2698 v4 @ 2.20GHz type (looping experiments) or on another machine with 4 NVIDIA GTX 1080 GPUs with 12GB memory, 692GB RAM and 72 cores of Intel(R) Xeon(R) Gold 6140 CPU @ 2.30GHz type (only training).
\subsection{Random Instantiation}
\begin{figure}
  \centering
  \includegraphics[width=0.35\textwidth]{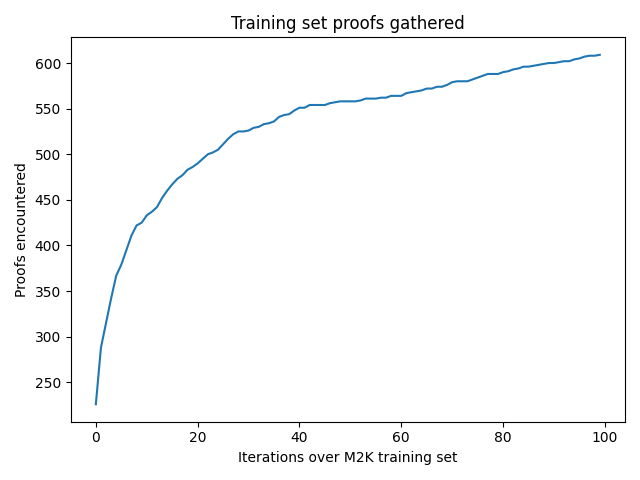}
    \caption{Cumulative M2k solutions by random instantiation limited to the training set.}
    \label{fig:M2k-rs}
\end{figure}

In Figure~\ref{fig:M2k-rs}, we show the cumulative amount of training M2K problems
solved by 100 runs of the random instantation process
(Section~\ref{sec:random}) with the 25-5 sampling setting. New problems
are solved, but after a while, the process plateaus at just above 600
M2K training set problems solved. Note that the random instantiation
process only tries constants in the second pass (Section~\ref{sec:random}), thus grounding all clauses. The neural network
must instead learn by itself to ignore the non-constants when it needs to ground
a term.


\subsection{Training Procedure}
\label{training_procedure}
Here, we give more details about the training of the neural network
and the setup used.  The initial training data were the cumulative
proofs obtained by 9 runs of the random instantiator with 25 samples
on $level_0$ and 5 on $level_1$. If multiple proofs were found for a
problem, one of those proofs was chosen randomly. If a given proof
needed two levels of instantiation, the proof is split into 2 training examples $E_1, E_2$.
The input part of $E_1$ is the original CNF problem, while for $E_2$
it is the CNF problem with the right head symbols (and corresponding
fresh variables) filled in for the proof-related clauses. For $E_1$, the label part corresponds to the head symbols in the input of $E_2$, while the labels for $E_2$ are the constants that ground the terms.

\paragraph{\textbf{Label Construction:}} We concatenate all the proof
instantiations for a clause into a single sequence. The RNN component
was trained to predict this concatenated sequence of instantiations so
that the model can capture the conditional dependence between multiple
instantiations of one clause. To make it possible for the model to stop
instantiating a clause, a special \textit{stop} vector is added in
addition to the actual symbol vectors when comparing with the symbol
request vectors. The label corresponding to the \textit{stop} vector
was added to the end of each label sequence. For clauses where no
instantiation is part of the proof, the label sequence consists only
of the \textit{stop} label.

\paragraph{\textbf{Instance Shuffling:}} While we choose to handle multiple instances for a clause sequentially with an RNN, there is no ordering on these instances for a given clause. Therefore, during training, we randomize the order in which the different label sequences corresponding to different instances are concatenated. In principle, the same holds for the ordering on the variables, but this ordering is not randomized in the current setup. The variable order is defined by the order they appear in the clause. 

\paragraph{\textbf{Loss Balancing:}} The number of choice points, and thus the number of contributions to the total loss when naively added, is not the same for each training example. Some training examples require as many as 100 symbols to be chosen, while others need less than 10. Therefore, we normalize the loss contribution coming from each training example in the batch by the amount of choice points (i.e. the total length of all the concatenated label sequences for all clauses with variables in the training example). We then minimize the sum of these averages. The loss now corresponds more to our usecase: for example it is more important to get all 3 instances for a small problem, than it is to get 3 out of 80 instances for another bigger problem.

\paragraph{\textbf{Hyperparameter Settings:}} The GNN was used with node and layer dimensions, for all sets of nodes, set to 64. The network uses 10 message passing steps, with different layer parameters at each step. The RNN consists of a linear neural network layer with 3*64 inputs and 64 output dimensions followed by a rectified linear unit activation function, followed by a linear layer of input size 64 and output size of 64. To optimize the parameters of the network, the ADAM algorithm was used with \textit{learning rate} $0.0001$, minimizing the cross entropy between the symbols used in the known proofs and the predicted symbols. The maximum number of RNN iterations per input clause (which limits the total number of symbols chosen per clause) was set to 12. When auto-regressively sampling, we used a temperature parameter of $2$. This value was determined by a small search on the validation set (the values 1, 2, 3, 5 and 8 were tried).

\subsection{Training Curves}
\begin{figure}
    \centering
    \includegraphics[width=0.35\textwidth]{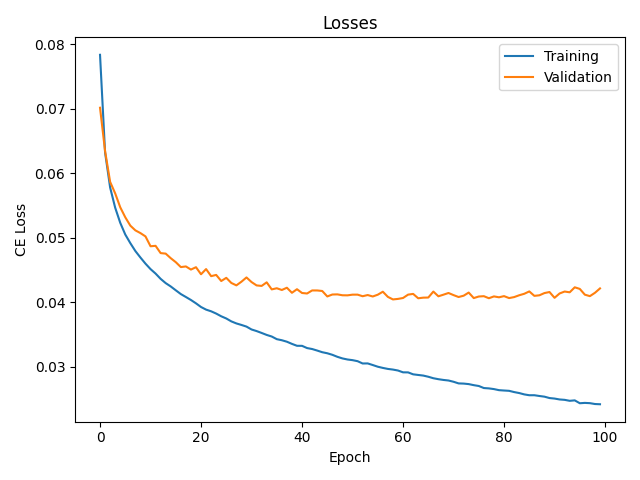}
    \caption{Loss curve for model trained on the 25-5 random instantiator data.}
    \label{fig:loss_curve}
\end{figure}

In Figure~\ref{fig:loss_curve}, the training and validation loss curves are shown for a model training on the proofs found by random instantiator on the full training set. After 40 epochs, the improvement of the validation loss slows down. In Figure~\ref{fig:medacc}, we show the median accuracy. We calculate whether the highest softmax output corresponds with the labels, and calculate the accuracy for each problem. We then take the median, as an indication of how well we make the right decisions. There, again 40 epochs are enough for most of the improvement. For the experiments in Section \ref{Fullloop}, we use checkpoint 79 because it is the earliest model with the highest median validation accuracy seen.
In the M2K experiments, such as in Section \ref{M2kloop}, we use the 56th checkpoint, for the same reason. 
\subsection{Evaluation of instance coverage per level}
\begin{figure}[b]
    \centering
    \includegraphics[width=0.35\textwidth]{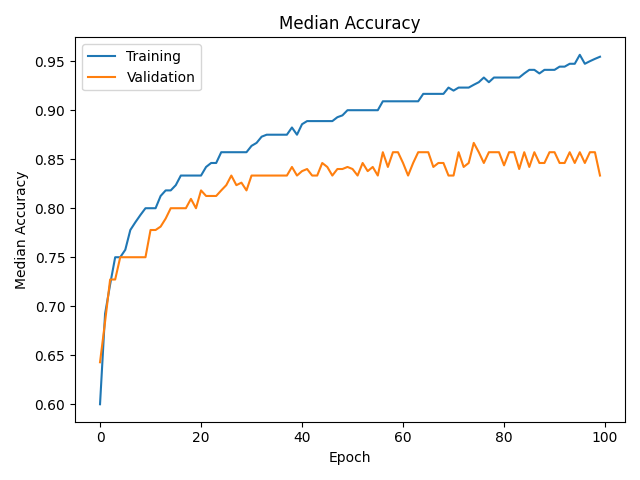}
    \caption{Median accuracy for model trained on the 25-5 random instantiator data.}
    \label{fig:medacc}
\end{figure}

While the validation accuracy seen in Figure~\ref{fig:medacc} is promising, the shown setting does not fully reflect how the model is used for the real instantiation task. There, the accuracy is computed while the input and the previous choices are always fully correct (i.e.\ according to the labels). However, in the practical setting, the model can only run autoregressively on its own output. For this setting, data is shown in Table~\ref{tab:table_info}. There we see the fraction of label instances covered as a function of the number of samples taken per clause (more simply, the accuracy). The accuracy values corresponding to certain quantiles \textit{q} are shown. For example, a $0.67$ score for $q=0.1$ means that for the worst 10\% of problems, at least 33\% of the required instances are still missing. While the median ($q=0.5$) and the 90th quantile indicate that with 25 samples, most instances are covered, the $q=0.1$ data show that there is a significant portion of problems for which instances are missing \todo{is there a smoother way to present this data?}. From this point, we always use 25 samples on $level_0$ and 5 samples on $level_1$.\todo{this last sentence could be more clearly separated from the analysis before it}
\begin{table*}[h]
  \begin{small}
    \centering
   \begin{tabular}{lrrrrrrrrr}
\toprule
{Number of samples per clause} &     1 &     2 &     3 &     5 &     7 &    10 &    15 &    20 &    25 \\
\midrule
$level_0$ - q=0.1 &  0.08 &  0.23 &  0.30 &  0.40 &  0.50 &  0.53 &  0.63 &  0.67 &  0.67 \\
$level_0$ - q=0.5 &  0.50 &  0.67 &  0.75 &  0.83 &  0.86 &  1.00 &  1.00 &  1.00 &  1.00 \\
$level_0$ - q=0.9 &  1.00 &  1.00 &  1.00 &  1.00 &  1.00 &  1.00 &  1.00 &  1.00 &  1.00 \\
$level_1$ - q=0.1 &  0.00 &  0.00 &  0.00 &  0.21 &  0.42 &  0.50 &  0.50 &  0.67 &  0.82 \\
$level_1$ - q=0.5 &  1.00 &  1.00 &  1.00 &  1.00 &  1.00 &  1.00 &  1.00 &  1.00 &  1.00 \\
$level_1$ - q=0.9 &  1.00 &  1.00 &  1.00 &  1.00 &  1.00 &  1.00 &  1.00 &  1.00 &  1.00 \\
\bottomrule
\end{tabular}
    \caption{Coverage of the instantiations needed for the proof, on problems from the validation set that the random instantiator found a proof for. The accuracy corresponding the quantiles 0.1, 0.5 and 0.9 are given. Note that even if the instantiations of the proof in our validation set for a given problem are not covered, the predicted instances might constitute a different proof.}
    \label{tab:table_info}
  \end{small}
\end{table*}
\subsection{Self-Improving Loop (M2K Dataset)}
\label{M2kloop}
While the previous section indicates that the model has learned to
recreate the right instances for many proofs on unseen data, in order
to be truly useful, the system needs to be able to generate new
proofs, and then learn from these new proofs, in a self-improving
loop. To test the setup for this capability, we do a looping
experiment. The model is trained until the validation accuracy stops
improving on the seed data generated by the random instantiatior, then
we predict on problems from the training set, including those where the
random instantiator could not find a proof. We then keep all the
proofs, and train again including the new proofs.
\begin{figure}
    \centering
    \includegraphics[width=0.35\textwidth]{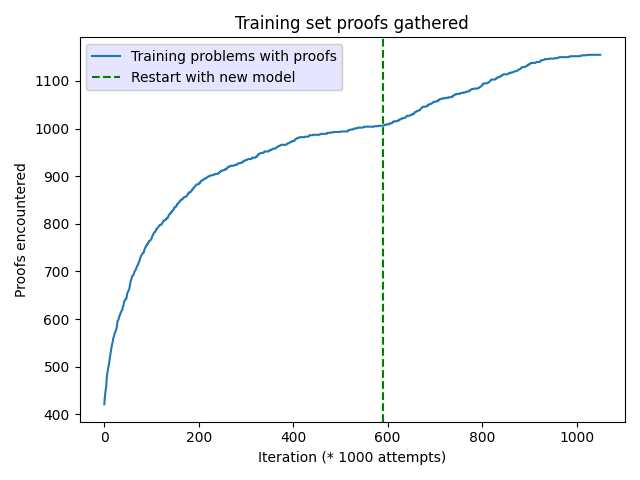}
    \caption{Number of problems for which a solution was found (cumulative, M2K)}
    \label{fig:cumulsolution}
\end{figure}

Because one full loop iteration can take 
time, we first test
this setup on the M2K subset. There are 4163 problems derived from the
M2K theorems assigned to the training set. Of these, the random
instantiator solved 421 total in 9 runs. This is used to train a model
(see Section \ref{training_procedure}). After this initial training,
we start the self-improvement loop. In each iteration, 1000 problems are
attempted and 1000 random previous proofs are trained on (but the model
parameters are kept between iteration). Every 10 iterations, we also
run the proof attempts on the test set.

\begin{figure}[b!]
    \centering
    \includegraphics[width=0.35\textwidth]{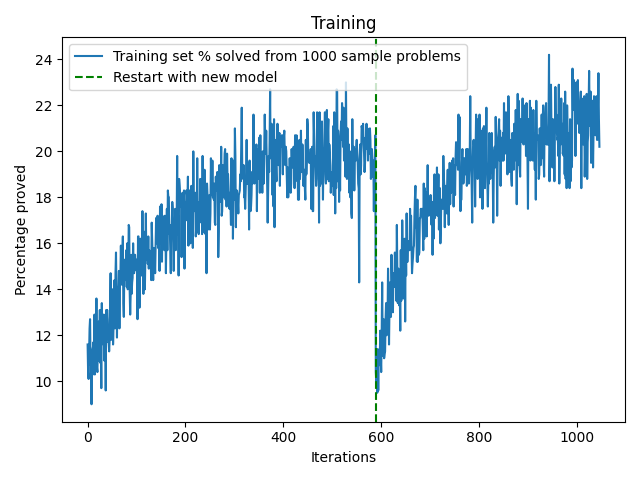}
    \caption{Percentage of problems solved (training set problems, M2K)}
    \label{fig:trainprov}
\end{figure}
Starting with 421 initial problems, after 590 loop iterations, there are 1006 problems from the training set that we have a proof for (see Figure \ref{fig:cumulsolution}). However, the discovery of new proofs stagnates after around 400 iterations. We concluded that restarting the training with a fresh model might bring more new proofs, as by this point the training solutions had been seen many times, which could lead to an overfit model. As seen in Figure~\ref{fig:cumulsolution}, this restarts the process and the system finds 149 more proofs, for a total of 1155.

\begin{figure}[b!]
    \centering
    \includegraphics[width=0.35\textwidth]{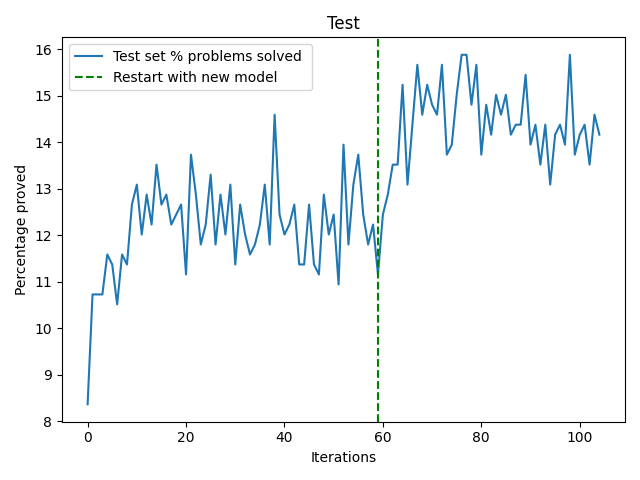}
    \caption{Percentage of problems solved (test set problems, M2K)}
    \label{fig:testprov}
\end{figure}

In Figures~\ref{fig:trainprov} and~\ref{fig:testprov}, the train and test performance of the system are shown. The system learns how to reprove around 22\% of training problems, which corresponds roughly to the ratio between 1155 and 4163. On the test set, the system can prove 14.5\%, which is more than double the performance of 1 run of the random instantiator with the same sampling settings (6.9\%). Especially noteworthy is the extra performance gain after the restart. This indicates that the optimization process found a more generalizable optimum when trained from scratch with the solutions found before the restart of the self-improvement loop.

\subsection{Self-Improving Loop (Full Dataset)}
As the self-improvement loop was successful on the M2K dataset, the experiment was repeated for the larger, full dataset. All settings are the same, except there are 10.000 proof attempts and 10.000 training samples each iteration. While fewer iterations were possible due to the increased computational requirements, there was still observable improvement. The number of training problems for which a proof was known increased from 6592 to 10543 in 42 rounds of 10.000 attempts. As the full training set is 96532 problems, this corresponds to fewer than 5 full sweeps of the data. The random instantiator reached 9923 problems in 100 run over the full training set. We conclude that the system finds more proofs, faster than the random instantiator. As for the test set solving performance, this improved to $10.59\%$, while 1 run of the random baseline proved $3.3\%$.


\label{Fullloop}
\section{Related Work}
In addition to the recent general work on synthesis of logical data
mentioned in Section~\ref{sec:intro}, there is recent work on choosing
instantiations in automated reasoning. Several examples are found in the SMT community, where
gradient boosted tree algorithms were used to filter possible terms
\cite{blanchette:hal-02381430} and to rank them for the SMT solving
procedure \cite{DBLP:conf/sat/JanotaPP22}. These approaches however
are working within the solving loop of an existing SMT solver, whereas
we are synthesizing instances outside of the SMT procedure.

There is also work on synthesizing loop invariants, which is similar
in spirit to what is attempted in this work \cite{si2018learning}. A
difference to their approach is that we are synthesizing many objects
(instances of each clauses) simultaneously, whereas loop invariant
synthesis is more concerned with a single object. Also, the specific
grammar of loop invariants used is limited, whereas we jointly learn
synthesis over arbitrary function signatures, within a single
signature-invariant system.


\section{Conclusion \& Future Work}
We have shown that a fully neural instantiation mechanism for many clauses at the same time is feasible. Starting from data generated by randomly instantiating variables in problems from a real-world mathematics dataset, the machine learning component can learn how to instantiate and improve convincingly better than the random component. 

\todo{add percentages}

The combination of a neural instantiator with a strong ground solver with a congruence closure mechanism combines two techniques according to their respective strengths: the global heuristics that the graph neural networks can learn for instantiating the first-order variables are combined with the fast and optimized reasoning components of SAT-based solvers to propagate the consequences of the instantiations.

Currently, a limitation is the growth of memory and time consumption as a function of inaccurate instantiations. The extra clauses generated slow down the process, so that running the process in an iterative manner is costly. Therefore, a premise selection mechanism could be included in between the instantiation phases in the future. Instantiations that are useless are nevertheless generated by sampling from the predictor's distribution. Much deeper proof terms could be generated if these could be efficiently pruned.
\clearpage
\bibliography{mj,references}

\end{document}